\documentclass[conference]{IEEEtran}
\IEEEoverridecommandlockouts
\usepackage{cite}
\usepackage{amsmath,amssymb,amsfonts}
\usepackage{algorithmic}
\usepackage{graphicx}
\usepackage{textcomp}
\usepackage{xcolor}
\usepackage{booktabs}
\usepackage{multicol}
\usepackage{multirow}
\usepackage{subcaption}
\usepackage{hyperref} 
\usepackage{xurl} 
\def\BibTeX{{\rm B\kern-.05em{\sc i\kern-.025em b}\kern-.08em
    T\kern-.1667em\lower.7ex\hbox{E}\kern-.125emX}}
\begin{document}
\title{Sometin Beta Pass Notin: Improving Multilingual ASR for Nigerian Languages via Knowledge Distillation
}


\author{\IEEEauthorblockN{Sewade Ogun}
\IEEEauthorblockA{
\textit{Masakhane NLP}\\
0009-0003-0648-021X}
}

\maketitle

\begin{abstract}
Although modern multilingual Automatic Speech Recognition (ASR) systems support several Nigerian languages, their performance consistently lags behind resource-rich languages such as English and French. Nigerian languages present unique modelling hurdles, including acute data scarcity, inconsistent orthography, tonal diacritics, diverse accents, frequent code-switching, and localised named entities. To address these challenges, we developed a multilingual ASR framework using a two-stage distillation process. First, we employed student-teacher knowledge distillation from existing monolingual models, conditioned on robust language-specific N-gram language models. Second, we performed iterative self improvement using pseudo-labelled data to further refine accuracy.
Our method significantly bridges the performance gap, achieving on average a reduction in the relative Word Error Rate (WER) of 29~\% over the monolingual baselines. Our models also outperform state-of-the-art  multilingual models across major benchmarks, including Common Voice and FLEURS. We introduce Sometin Beta Pass Notin (SBPN), a multilingual foundational ASR model that covers Yorùbá, Hausa, Igbo, Nigerian Pidgin, and Nigerian English. 
\end{abstract}


\section{Introduction}
\label{sec:intro}

Nigeria is home to over 500 distinct languages.
Many natives are inherently bilingual or multilingual, acquiring one or more indigenous languages through socialisation while attaining proficiency in the official language, English, through formal education. However, in certain regions, the Nigerian Pidgin has increasingly replaced indigenous mother tongues as the primary language of daily interaction  \cite{osoba2016language}. This transition places several languages at significant risk of linguistic attrition or extinction, which is further accelerated by lack of digital exposure. Resource-rich languages dominate the internet, creating a self-reinforcing cycle that further sidelines indigenous tongues.

Many Nigerian languages still have millions of speakers around the world. The languages with the highest number of speakers include Nigerian English (178~M) \cite{unuabonah2022introducing},\footnotemark\ Nigerian Pidgin (80.2~M), Hausa (63~M), Yorùbá (45.7~M), and Igbo (33~M). Yet, they remain classified as low-resource languages due to several critical factors \cite{nigatu-etal-2024-zenos}. Primarily, this is due to the scarcity of large-scale high-quality Automatic Speech Recognition (ASR) datasets. Although recent community initiatives have produced open-source datasets and baseline models for Yorùbá, Hausa, Igbo \cite{emezue25_interspeech}, Nigerian Pidgin \cite{rufai2020towards} and Nigerian English \cite{olatunji-etal-2023-afrispeech}, these speech-text pairs consist largely of read speech based on predefined templates with limited textual diversity. Hence, these efforts often do not translate into better accuracy in real-life conversational or spontaneous testing scenarios \cite{furui2005recognition}.
A further challenge is the prevalence of code-switching in conversational speech. For example, numerical data, such as dates, measurements, and years, are typically spoken in English during everyday interactions, even when the primary language of communication is different. Furthermore, tonal languages like Yorùbá and Igbo rely on diacritical marks to convey correct pronunciation and meaning; any omissions or inconsistencies in these marks within training data lead to increased modelling errors. 
Lastly, modelling loanwords presents a distinct hurdle \cite{ikotun2023cross}. The etymology of these languages reveals a significant cross-linguistic influence. Specifically, Nigerian languages have adopted numerous English words often by modifying their orthography or pronunciation.\footnotetext{Nigerian English is a distinct variety of English with a unique accent and a different semantics (e.g., ``to flash'' for a missed call) that distinguish it from Standard English}

While open-source monolingual ASR models exist for Nigerian languages, we hypothesize that combining them into a unified multilingual architecture can mitigate language-specific data limitations and yield stronger joint acoustic representations. Given the scarcity of annotated target speech, we employ pseudo-labelling to adapt the model to real-world conversational dynamics.  
The following are our contributions.
\begin{itemize}
\item We introduce the first  multilingual foundational ASR model, Sometin Beta Pass Notin (SBPN),\footnote{Sometin Beta Pass Notin is a Nigerian Pidgin expression meaning ``Something is better than nothing''. It follows the theme that learning from several average teachers can still produce a good student.} that focuses on Nigerian languages. The model comes in two variants, SBPN-Base\footnote{\url{https://huggingface.co/ogunlao/SBPN_multilingual_base}} and SBPN-Large.\footnote{\url{https://huggingface.co/ogunlao/SBPN_multilingual_large}}
\item We develop N-gram language models for 5 Nigerian languages, and then demonstrate their use in improving ASR performance.
\item We show the use of large scale pseudo-labelling for improving low-resourced Nigerian languages.
\end{itemize}
The report is divided into the following sections. Section~\ref{sec:literature_review} examines the existing literature on ASR knowledge distillation, including some related multilingual ASR research on African languages. Section~\ref{sec:methodology} describes our methodology in detail while Section~\ref{sec:experiments} details the experiments and results. We discuss some future directions and conclude in Section~\ref{sec:future}.

\section{Related Work}
\label{sec:literature_review}
Training ASR models with pseudo-labelled data is a widely explored area in the literature  \cite{9846970, 10225353, bhogale2024empowering}. On a large scale, this has yielded significant improvements in monolingual and multilingual foundational ASR models  \cite{radford2023robust, gandhi2023distil, sekoyan2025canary}.  
The pseudo-labels are often improved using methods like shallow fusion of the ASR predictions with a strong external language model in the target domain during inference, confidence filtering of the generated pseudo-labels \cite{9054295}, pseudo-label refinement using generative error correction  \cite{yang2024large}, etc. In low-resource scenarios, a well trained acoustic model is typically finetuned on the small amount of curated dataset. Then, pseudo-labels are generated for larger unlabelled data using the finetuned model \cite{xu2021self}. 

Recent projects on ASR for African languages include the development of the AfriHUBERT model \cite{alabi25_interspeech}, a project that extended the mHuBERT-147 model from 16 African languages to 1{,}226 African languages, including the Igbo, Yorùbá, and Hausa languages. However, the performance reported on the selected Nigerian languages still lags behind others. Therefore, in this work, we explore ways to improve the accuracy of speech recognition models on Nigerian languages.

\section{Methodology}
\label{sec:methodology}
\subsection{Dataset curation and preprocessing}
The proposed methodology relies on curated labelled and unlabelled speech datasets for target Nigerian languages, Yorùbá (yo), Nigerian Pidgin (pcm), Hausa (ha), and Igbo (ig), and Nigerian English (en-ng). The data were sourced from online speech dataset repositories, audio-books, online digital sources such as radio shows, online audio platforms, and freely available podcasts with permissive licenses. Rather than relying on pseudo-labelled data for English, we supplemented the Nigerian English subset with a larger, publicly available Standard English (en) corpus with labels. Adding this broader set of speakers and accents prevents the ASR model from overfitting to the limited sample size of the target corpus. Table~\ref{tab:5 langugages} lists the labelled dataset, the language(s) covered, and the corresponding hours of training data available.

\begin{figure}[t] \centering{
\includegraphics[width=0.98\linewidth]{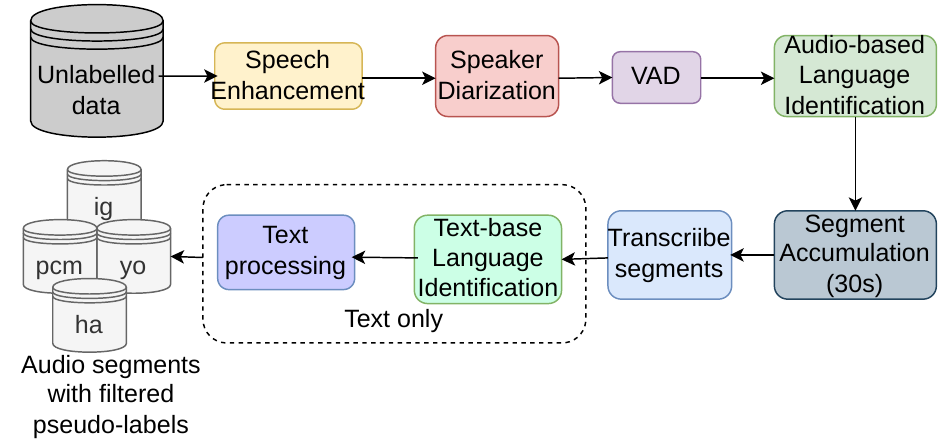}}
\caption{Flow diagram showing the pseudo-label generation pipeline from unprocessed recordings to processed audio segments with pseudo-labels.}
\label{fig:pseudolabel_pipeline}
\end{figure}

\begin{table}[t]
\caption {Curated ground-truth labelled datasets used for training SBPN including the languages in each dataset and the total number of hours of the training subsets.} \label{tab:5 langugages}
\centering
\begin{tabular}{ l  c  c }
\toprule
\textbf{Dataset} & Languages & Duration (h)\\
\midrule
Common Voice  \cite{commonvoice:2020} & yo, ha, ig, en-ng & $13.5$ \\
Naijavoice dataset   \cite{emezue25_interspeech}& yo, ha, ig & $1{,}862.0$ \\
FLEURS dataset  \cite{10023141} & yo, ha, ig & $37.4$ \\
SLR86  \cite{gutkin-et-al-yoruba2020}& yo & $4.0$ \\
BibleTTS  \cite{meyer22c_interspeech} & yo, ha & $111.7$ \\
Igbo-asr\footnote{https://www.kaggle.com/code/jameskaile/igbo-asr/input} & ig & $23.2$ \\
Nigerian pidgin dataset  \cite{rufai2020towards} & pcm &  $5.4$ \\
Afrispeech-200  \cite{olatunji-etal-2023-afrispeech} & en-ng & $172.4$ \\
Gigaspeech-L  \cite{chen2021gigaspeech} & en (standard) & $2{,}483.9$ \\
\bottomrule
Total & - & $4{,}713.5$ \\
\bottomrule
\end{tabular}

\end{table}
Unlike the labelled data, the unlabelled speech data in their original form were multi-speaker and varied in recording quality and length. Therefore, the data required post-processing for ASR model training. The entire data processing pipeline is depicted in Fig.~\ref{fig:pseudolabel_pipeline}. We adopted the audio processing pipeline described in \cite{kimiteam2025kimiaudiotechnicalreport}. Concretely, each recording was denoised using a speech enhancement model, MossFormer2 \cite{zhao2023mossformer},\footnotemark\ and then split into individual speaker segments using the Pyannote speaker diarization toolkit  \cite{Bredin23}. The diarization process groups segments into individual speakers. Segments with speaker embedding similarity exceeding $0.7$ were also combined into the same speaker group to produce longer single-speaker continuous speech as proposed in \cite{kimiteam2025kimiaudiotechnicalreport}.\footnotetext{Initial pseudo-labelling with original samples yielded bad pseudo-labels, so we denoised the noisy samples.} 
The audio segments were further passed through a voice activity detector (VAD)\footnote{\url{https://github.com/snakers4/silero-vad}} to remove silence segments and other non-speech segments. The VAD outputs were subsequently post-processed by keeping the silence segments that are less than 1.5~s in-between the VAD processed segments. Additionally, we filtered out segments that did not belong in our target languages using a two step filtering process. The language of each audio segment is first predicted using an audio-based language identifier (LID) trained on the VoxLingua107 dataset  \cite{valk2021slt}.\footnote{\url{https://huggingface.co/speechbrain/lang-id-voxlingua107-ecapa}} The LID model can identify 107 languages including Yorùbá, English and Hausa, but was not trained to identify Nigerian Pidgin and Igbo languages. For these remaining languages, only a second-step text-based language identification with filtering was performed using AfroLID \cite{adebara2022afrolid}. Given that the removed audio comprised less than 1 \% of the total corpus, no ablation studies were conducted to isolate the effect of these removals.

Finally, all long segments were divided into 30~s segments using a silence threshold of $-50~dB$. This ensures that the segments are compatible with ASR models used for pseudo-labelling. The total number of hours after processing was about 10{,}000~h.

\subsection{N-gram model training and data pseudo-labelling}
\label{sec: n-gram-lm}
We identified existing open-source monolingual ASR models trained on each Nigerian language for initial pseudo-labelling of unlabelled data.\footnote{\url{https://hf.co/CLEAR-Global/w2v-bert-2.0-hausa_579_993h_yourtts}}$^{,}$\footnote{\url{https://hf.co/CLEAR-Global/w2v-bert-2.0-igbo_naijavoices_500h}}$^{,}$\footnote{\url{https://hf.co/CLEAR-Global/w2v-bert-2.0-yoruba_naijavoices_500h}}$^{,}$\footnote{\url{https://hf.co/asr-nigerian-pidgin/pidgin-wav2vec2-xlsr53}} 
All candidates employ the Connectionist Temporal Classification (CTC) objective \cite{10.1145/1143844.1143891}, which benefits from external language model (LM) fusion during inference.
\begin{figure}[t]
    \centering
    \includegraphics[width=1.0\linewidth]{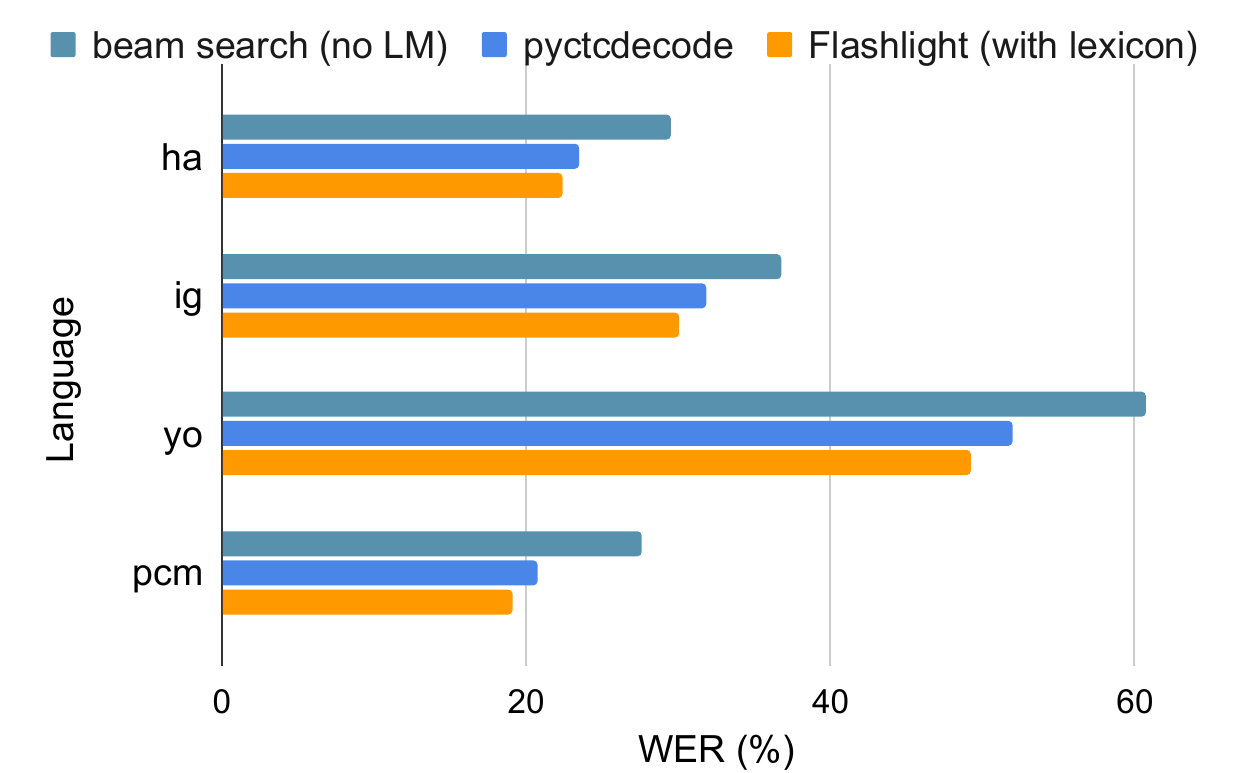}
    \caption{Comparing WER (\%) on the validation sets of Hausa (ha), Igbo (ig), Yorùbá (yo), and Nigerian Pidgin (pcm) when different CTC decoder libraries are used. (ha, ig, and yo were evaluated on FLEURS while pcm was evaluated on the Nigerian Pidgin validation set.)}
    \label{fig:lm_library_comparism}
\end{figure}
Hence, we develop an N-gram LM for each language using text corpora curated from training dataset labels in Table~\ref{tab:5 langugages}, Common-Crawl-100,\footnote{See \url{https://data.statmt.org/cc-100/}} and other language-specific text repositories.\footnote{See \url{https://github.com/angeloobeta/Igbo-datasets}}$^{,}$\footnote{See \url{ https://github.com/Niger-Volta-LTI/yoruba-text}}$^{,}$\footnote{See \url{https://github.com/ijdutse/hausa-corpus}} For Pidgin English, we selected the subset of the CLaT dataset \cite{lin23e_interspeech} and the NaijaSenti dataset \cite{muhammad-etal-2022-naijasenti} that contains Nigerian Pidgin. Due to the limited size of the Nigerian Pidgin corpus, we added the Nigerian English subset of the International Corpus of English (ICE).\footnote{See \url{https://varieng.helsinki.fi/CoRD/corpora/ICE-NIG/}}\ The ICE dataset contains texts extracted from Nigerian media and newspapers, and therefore provided more in-domain Nigerian phrases and words, including named entities.

Additional processing for the curated texts in each language includes text deduplication and filtering out sentences identical to those in the labelled data's validation and test sets after lower-casing and removing accents and punctuation from the sentences. For Yorùbá language, all texts without diacritical marks were removed. Using the filtered text corpus, we trained a 5-gram LM for each language with the KenLM library. The perplexities of the trained N-gram models evaluated on the FLEURS validation sets and the Nigerian Pidgin validation set are provided in Table~{\ref{tab:n_gram_table}}. The perplexity scores give some indications about the performance of the N-gram LMs. The high perplexity scores observed highlight the limited size of the text corpus, resulting in severe data sparsity and high frequency of out-of-vocabulary words. 
The lower perplexity for Nigerian Pidgin likely indicates a less diverse vocabulary in its corpus.

Subsequently, the LMs were integrated into the monolingual ASR models via shallow fusion to predict pseudo-labels for the unlabelled dataset. We considered only hard pseudo-labelled targets; taking the text sequence with the highest conditional probability given the speech sample. We compared different decoding libraries and selected the Flashlight CTC decoder \cite{kahn2022flashlight} because it provided better pseudo-labels for our validation sets as indicated in Fig.~\ref{fig:lm_library_comparism}. 
\begin{table}[t]
\centering
\caption{Table showing the perplexity of the 5-gram language models trained for Yorùbá (yo), Hausa (ha), Igbo (ig), and Nigerian Pidgin (pcm) using the KenLM library.}
\label{tab:n_gram_table}
\begin{tabular}{| l | c | c | c | c |}
\hline
Language & yo & ha & pcm & ig \\
\hline
Perplexity & 775.60 & 243.24 & 15.48 & 472.98 \\
\hline
\end{tabular}
\end{table}

\paragraph*{Nigerian pidgin orthographic normalization}
\label{special_pseudolabel}
Nigerian Pidgin lacks a standardized orthography; words frequently appear in plain English, spoken Pidgin, or written Pidgin variations \cite{adelani2025does}.
To improve target sequence consistency under data scarcity, we normalized contextual homophones across both the labelled and pseudo-labelled transcripts into single canonical tokens based on corpus word frequencies. Candidate sets were constructed by generating pseudo-labels for the Nigerian Pidgin training and validation splits using both an English ASR model \cite{sekoyan2025canary} and
a Nigerian Pidgin monolingual model. We aligned these predictions with the original transcripts via positional word clustering, taking a 4-word context window around each target token into account.
For instance, among the variants ``they", ``de", and ``dey" (with frequencies of 501, 30, and 1005, respectively), all occurrences were mapped to the most frequent variant, ``dey". While this clustering effectively grouped valid variants (e.g., \{``boi", ``boy"\}, \{``demself", ``demsef", ``themself"\}), it occasionally generated false positives; these were readily filtered out by discarding low-frequency candidates below a heuristic threshold.

In addition, another set of larger distinct homophones was revealed by the clustering. Some homophones cannot be interchanged contextually (e.g., ``say" vs. ``sey", or ``niger" vs. ``naija"). To resolve these context-dependent ambiguities, we queried the Nigerian Pidgin N-gram language model to select the homophone candidate that yields the highest sequence probability at the target position. Although the normalization process relied partially on manual intervention, future work will automate this process to generalize to other low-resource languages. Test set transcripts were left unnormalized to evaluate real-world generalization.

\subsection{Model architecture and knowledge distillation}
\label{sec:knowledge_distillation}
Using the labelled and pseudo-labelled pairs of speech and text training data, we developed a set of multilingual ASR models, SBPN-Base and SBPN-Large.
Their model architecture is based on the recurrent-neural-network-based Transducer (RNNT) \cite{graves2012sequence}. The RNNT architecture consists of a Fast Conformer encoder \cite{rekesh2023fast}, a stateful LSTM-based prediction network, and a feed-forward joint network. An auxiliary convolutional neural network-based CTC head is attached to the encoder to regularise the encoder features, which is especially useful in the presence of noisy pseudo-labelled data. The differences between SBPN-Base and SBPN-Large are presented in Table~\ref{tab:hyperparameters}.

To provide sufficient capacity for cross-lingual transfer, we selected a 600M-parameter architecture for SBPN-Large. Its encoder was initialized with the pre-trained weights from Parakeet-TDT-600-V3 to accelerate training, while randomly initializing all remaining layers. The training framework consists of two sequential phases:
\begin{itemize}
    \item Knowledge distillation step: The model is initially trained on a combination of pseudo-labelled and ground-truth data, followed by refinement on ground-truth data alone.
    \item Self-Improvement step: Using the best checkpoint from the initial phase, pseudo-labels are generated iteratively for each language via shallow fusion with a target-specific N-gram language model. To balance dataset scale and transcript quality, candidate pseudo-labels are filtered using language-specific confidence thresholds. Furthermore, pseudo-labels assigned a language tag that disagrees with the target speech language are discarded to eliminate mis-classifications carried over from initial data processing. Final training proceeds on the combined set of filtered pseudo-labels and ground-truth data until convergence. We perform the self-improvement step for two iterations.
\end{itemize}

To evaluate parameter efficiency, we then trained a compact 120M parameter model (SBPN-Base). The encoder was initialized from the Parakeet-TDT CTC-110M model.\footnote{Checkpoint available at \url{https://hf.co/nvidia/parakeet-tdt_ctc-110m}}\ For this variant, hard pseudo-labels were generated directly by the trained SBPN-Large model. 
During the pseudo-label generation stage, if the known target language was not the top beam hypothesis, the highest-ranked hypothesis matching that language was selected when available. While this target-language forcing improved the quality of pseudo-labels, it was omitted during benchmark test set evaluation for fair comparison with prior work.
After minimal confidence filtering to discard low-probability transcriptions, the pseudo-labels along with their speech segment pairs were combined with the ground-truth labelled corpus to train the student model.

\subsection{Experimental setup}
\subsubsection{Training and implementation details}
Both models were trained using a multi-task training setup, where the total loss is computed as a weighted loss between the RNNT loss and the CTC loss of the encoder head. We adopted the RNNT loss, implemented as Graph RNNT loss \cite{10096679} in NeMO, to reduce memory consumption during loss computation. The models were trained using the NeMo toolkit \cite{kuchaiev2019nemo} with the AdamW optimizer using a linear warm-up of 2{,}500 steps and cosine annealing. Temperature sampling was applied during data loading to mitigate language imbalance. The key hyperparameters and data augmentation settings are detailed in Table~\ref{tab:hyperparameters}. 

Text preprocessing include converting digits to English spoken format \cite{zhang21ja_interspeech} and stripping punctuation except apostrophes and hyphens.
Additionally, a unified sentence-piece tokenization with 4{,}096 subword tokens was trained from the combination of all labelled training data texts. A language tag in $\{$\texttt{<|en|>}, \texttt{<|ig|>}, \texttt{<|yo|>}, \texttt{<|pd|>}, \texttt{<|ha|>}$\}$ was prepended to each text label when loading the data, to reduce cross-lingual interference. It also serves as the language label used for language identification (LID) during inference.
\begin{table}[t]
\centering
\small
\caption{Training hyperparameters and augmentation settings}
\label{tab:hyperparameters}
\resizebox{\columnwidth}{!}{%
\begin{tabular}{lcc}
\toprule
\textbf{Parameter / Setting} & \textbf{SBPN-Base} & \textbf{SBPN-Large} \\
\midrule
\multicolumn{3}{l}{\textit{Model Architecture}} \\
Encoder layers & 17 & 24 \\
Encoder hidden dimension & 512 & $1{,}024$ \\
Number of LSTMs & 1 & 2 \\
Attention heads & 8 & 8 \\
Vocabulary size (tokens) & $4{,}096$ & $4{,}096$ \\
Log-Mel bins & 80 & 120 \\
RNNT loss type & Graph RNNT & Graph RNNT \\
\midrule
\multicolumn{3}{l}{\textit{Optimization \& Training}} \\
Base learning rate & $1\times10^{-4}$ & $3\times10^{-4}$ \\
Self-Improvement learning rate & $1\times10^{-5}$ & $1\times10^{-5}$ \\
Weight decay & $1\times10^{-4}$ & $1\times10^{-4}$ \\
CTC loss weight & 0.3 & 0.3 \\
Global batch size & 320 & 240 \\
Sampling temperature & 20 & 20 \\
Beam size & 100 & 100 \\
Checkpoint averaging & Last 3 & None \\
Float precision & BF16-mixed & BF16-mixed \\
\midrule
\multicolumn{3}{l}{\textit{Data Augmentation (Distillation Stage $\rightarrow$ Self-Improvement Stage)}} \\
SpecAugment & \multicolumn{2}{c}{Enabled} \\
Noise addition (MUSAN) \cite{snyder2015musan} & \multicolumn{2}{c}{$p=0.4 \rightarrow 0.25$ (SNR: 5--30\,dB)} \\
Time stretching ($\{0.9, 1.0, 1.1, 1.2\}$) & \multicolumn{2}{c}{$p=0.4 \rightarrow 0.25$} \\
\bottomrule
\end{tabular}%
}
\end{table}

\subsubsection{Conversational Speech Simulation via WSOLA}
\label{sec:wsola_data}
To evaluate robustness against the variable speaking rates characteristic of spontaneous speech (e.g., rapid enunciations and mid-utterance pauses), we simulated speed perturbations across target test sets. The speaking rate of the test samples was modified using a stretching factor that ranges from slow speech to twice the baseline speaking rate. To modify them while preserving the natural voice timbre, we applied the Waveform Similarity-Based Overlap-Add (WSOLA) algorithm \cite{roelands93_eurospeech}.\footnote{Implemented in \url{https://github.com/KAIST-MACLab/PyTSMod}}\

\subsubsection{Evaluation set and metrics}
SBPN models were evaluated primarily on the FLEURS test set and the Common Voice test sets for the Yorùbá, Hausa, and Igbo languages. Performance for Nigerian Pidgin and Nigerian English was evaluated using the Nigerian Pidgin test set and the Nigerian Common Voice test set respectively. 
The metrics used for the evaluations are the word error rate (WER) for transcription accuracy and the micro averaged $F_1$ score for language identification. In the language identification task, it is assumed that every test utterance in the test set belongs to the language indicated by the test set metadata.

\section{Experiments}
\label{sec:experiments}
\begin{table}[t]
\centering
\caption{WER (\%) on the FLEURS test sets and Nigerian Pidgin test set for monolingual teacher models and the SBPN-Large student models across knowledge distillation (KD) and self-improvement (SI) stages.}
\label{tab:table_main}
\begin{tabular}{l c c c c c c}\toprule
 & en-ng & ha & ig & yo & pcm & Average \\\midrule
Teachers & 25.3 & 31.04 & 38.68 & 55.6 & 32.44 & 36.61 \\
+ N-gram LM & - & 26.26 & 34.18 & 43.77 & 20.09 & 31.08 \\ \midrule
Training stages\\
Stage 1 (KD.) & 21.09 & 24.47 & 35.15 & 41.06 & 13.19 & 26.99 \\
Stage 2 (SI.) & \textbf{19.36} & \textbf{24.38} & \textbf{33.86} & \textbf{39.94} & \textbf{12.94} & \textbf{26.10} \\ \bottomrule

\end{tabular}
\end{table}

\subsection{Teacher vs. student performance}
\label{sec:average_tachers}
We evaluate the effectiveness of the two-stage distillation framework (Section~\ref{sec:knowledge_distillation}) by comparing the distilled SBPN-Large student model with the monolingual baseline teachers in the target languages. Table~\ref{tab:table_main} illustrates the primary findings of the experiments. The SBPN-Large model outperforms all the monolingual ASR teacher models on average by a large margin, a 29~\% relative reduction in WER over the baselines. Even with LM fusion with the baseline ASR models, the baseline performance lags behind SBPN-Large, which achieves a 16~\% relative reduction in WER. This demonstrates that the student model successfully absorbed and surpassed the knowledge distilled from the teacher. Examining the results of each stage, the distillation stage provided, on average, a significant relative reduction in WER of 26~\% over the monolingual teacher models. Although the WER was subsequently reduced further during the self-improvement stage, the reduction is smaller than in the first stage, indicating that most of the knowledge transfer occurred at the first stage. In our analysis, the second stage mainly helped improve performance on diacritics, code-switched data, and long speech, which are improvements not necessarily reflected in the test set results.

The highest improvement in ASR performance is on Nigerian Pidgin, boasting a relative reduction of 60~\% in WER over the teacher model. This could be attributed perhaps to the similarity of the Nigerian Pidgin language to Standard English, and also the text processing applied to the Nigerian Pidgin (pcm) pseudo-labels (Section~\ref{special_pseudolabel}). The improvement is also reflected in the Nigerian English (en-ng) results, specifically after the self-improvement stage, where we observe a relative WER reduction of 23~\% over a strong English baseline (Parakeet-TDT-600-V3). However, this reduction is not as significant as that of Nigerian Pidgin, perhaps due to the limited amount of Nigerian English data used for model training.\footnote{The entire AfriSpeech-200 training set is about 172~h and only a fraction of the accents is Nigerian.}\ The least improvement is observed in the Igbo language. Even with a large amount of pseudo-labelled data (1{,}900~h), we were only able to achieve a relative WER improvement of 13~\% over the baseline. Examining the curated audio recordings indicates that a large portion of Igbo speech involves code-switching between the Igbo language and the English language or the Nigerian Pidgin, which may have affected performance.

\subsection{Comparison with state-of-the-art multilingual ASR models.}
To assess whether parameter efficiency impacts performance relative to larger baselines, we evaluate both SBPN-Base (120M) and SBPN-Large (600M) against existing state-of-the-art (SOTA) multilingual models supporting Nigerian languages. Table~\ref{tab:several_asr_wer} reports WER on the FLEURS and Common Voice benchmark test sets. First, within their model size groups, SBPN-Base and SBPN-Large show significant improvements over existing multilingual baselines. SBPN-Base shows an average relative reduction in WER by 60~\% compared to AfriHuBERT and by 63~\% compared to mHuBERT-147 on the Common Voice test sets.
SBPN-Base even outperforms larger models on average on the FLEURS benchmarks, validating our focus on Nigerian languages.
In the billion parameter group, although SBPN-Large has only 600M parameters, a fraction of a billion parameters, it still outperforms all models on average on all the Nigerian language test sets. Compared to the next best model, SBPN-Large reduces the average WER on the FLEURS test sets by 21~\% relative to the MMS-1B multilingual model.

When comparing the results on the Common Voice test sets with FLEURS test sets, SBPN models show a higher reduction in WER on the Common Voice test sets than on the FLEURS test sets. Examining the test set labels indicates that the Common voice test sets are probably of higher quality in terms of orthography, diacritics, etc. For example, as pointed out by the authors of AfriHuBERT  \cite{alabi25_interspeech}, many texts in the Yorùbá FLEURS test sets do not have diacritical marks.

\begin{table}[]
\centering
\caption{Comparison with other SOTA multilingual models that support Nigerian languages. The large models were compared on the FLEURS test sets while the smaller models were compared on Common Voice test set to be consistent with reports in other works. We compare the results directly with values reported in other published works.}
\label{tab:several_asr_wer}
\begin{tabular}{l c | c c c c}\toprule
 & Model size & ha & ig & yo & Average \\\midrule
\multicolumn{6}{l}{Evaluated on the FLEURS test sets} \\
Whisper Large & 1.5~B  & 144.33 & 101.49 & 103.56 & 116.46 \\
MMS-1B & 1~B  & 25.51 & 44.61 & 53.56 & 41.23 \\
SeamlessM4T v2 & 2.3~B & - & 96.9 & 83.5 & 90.2 \\ \hline
SBPN-Base & 120~M  & 27.04 & 39.53 & 43.83 & 36.80 \\
SBPN-Large & 600~M & \textbf{24.38} & \textbf{33.86} & \textbf{39.94} & \textbf{32.72} \\ \hline
\multicolumn{6}{l}{Evaluated on the Common Voice test sets} \\
AfriHuBERT & 95~M  & 51.1 & 60.5 & 81.2 & 64.27 \\
mHuBERT-147 & 95~M  & 59.4 & 62.3 & 86.9 & 69.53 \\ \hline
SBPN-Base & 120~M  & 19.22 & 33.52 & 23.86 & \textbf{25.53} \\
SBPN-Large & 600~M  & \textbf{17.69} & \textbf{31.46} & \textbf{23.32} & \textbf{24.16} \\\bottomrule

\end{tabular}
\end{table}

\subsection{Robustness to conversational and spontaneous speech}
\begin{figure}[t]
    \centering
    \includegraphics[width=1\linewidth]{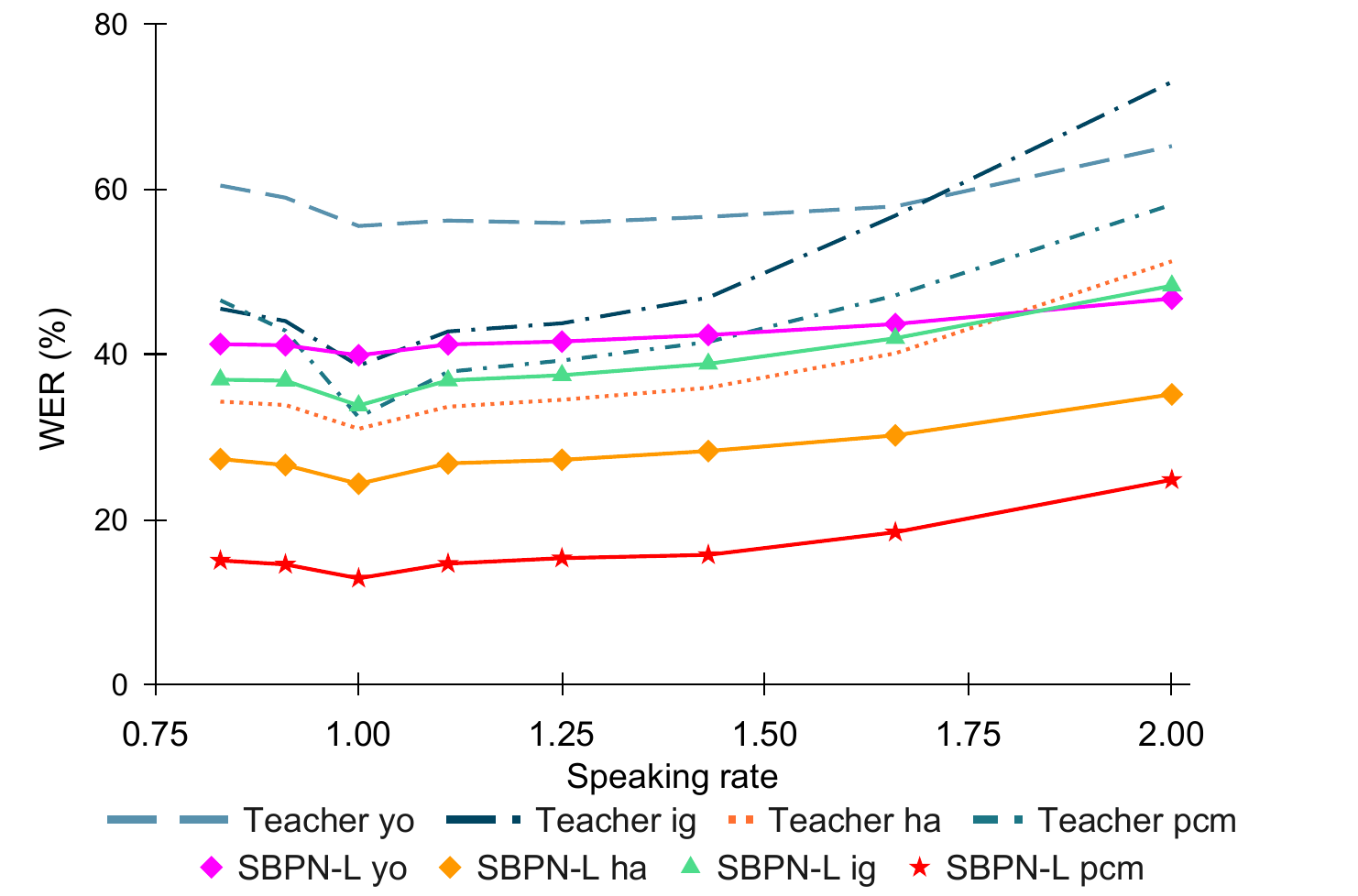}
    \caption{Performance of SBPN-Large on test set samples across several speaking rates (0.8x to 2x). Average WER (\%) computed on the FLEURS and Nigerian pidgin test sets.}
    \label{fig:sbpn_fastspeech}
\end{figure}
Conversational speech poses significant challenges to ASR systems due to irregular speaking rates and spontaneous disfluencies. Using the WSOLA-perturbed test sets (Section~\ref{sec:wsola_data}), we evaluated the impact of varying speech rates, ranging from severe time-stretching to 2$\times$ speed acceleration, on model accuracy.
As shown in Figure~\ref{fig:sbpn_fastspeech}, SBPN-Large maintains relatively consistent WERs against speed fluctuations, whereas baseline monolingual models show marked degradation at higher speaking rates. This indicates that SBPN-Large is better at transcribing fast conversational speech across Nigerian languages. Additionally, unlike the Igbo baseline model, which showed a sharp WER increase when the speaking rate was doubled, SBPN-Large remained relatively stable. Its WER increased by only 43~\% over the baseline speaking rate, compared to an 89~\% increase for the Igbo teacher model.

\subsection{Analysis of diacritical mark prediction}

\begin{figure}[t]
    \centering
    \begin{subfigure}[t]{0.49\columnwidth}
        \centering
        \includegraphics[width=\textwidth, height=1.0in, keepaspectratio]{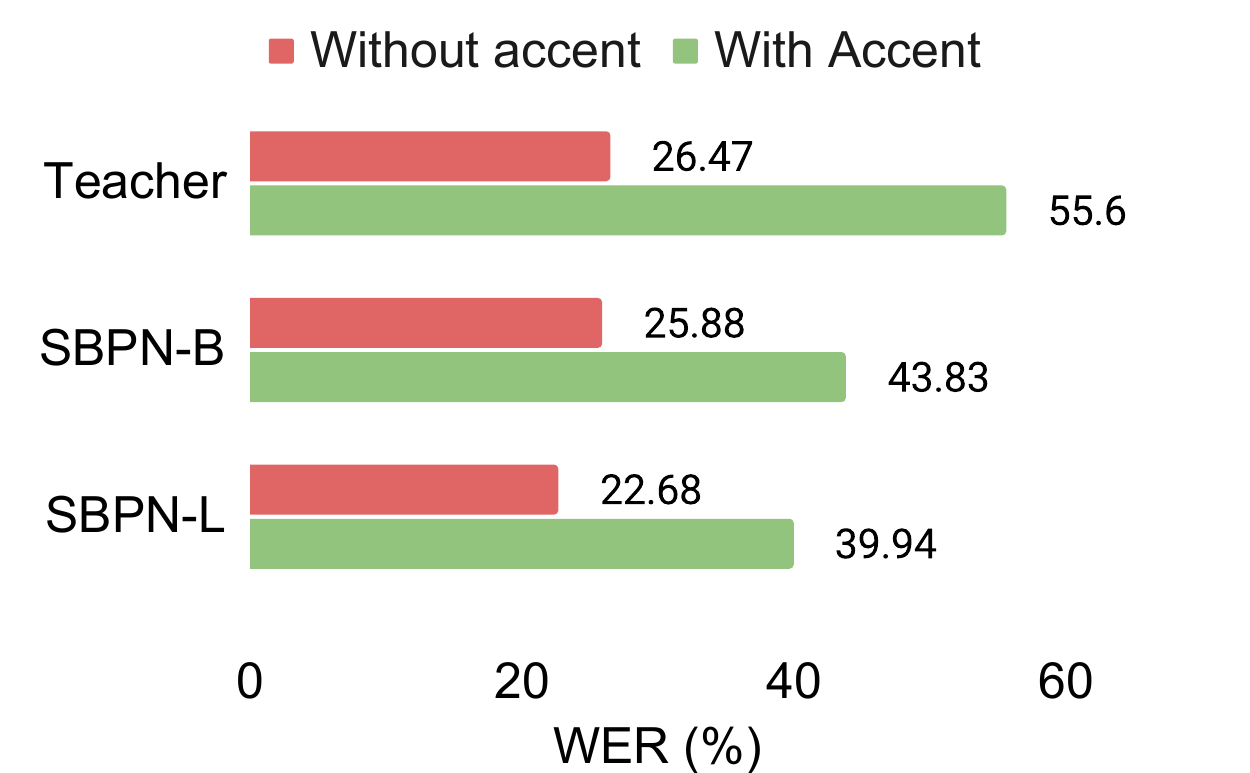}
        \caption{Yorùbá}
        \label{fig:accent_wer_yor}
    \end{subfigure}
    \hfill
    \begin{subfigure}[t]{0.49\columnwidth}
        \centering
        \includegraphics[width=\textwidth, height=1.0in, keepaspectratio]{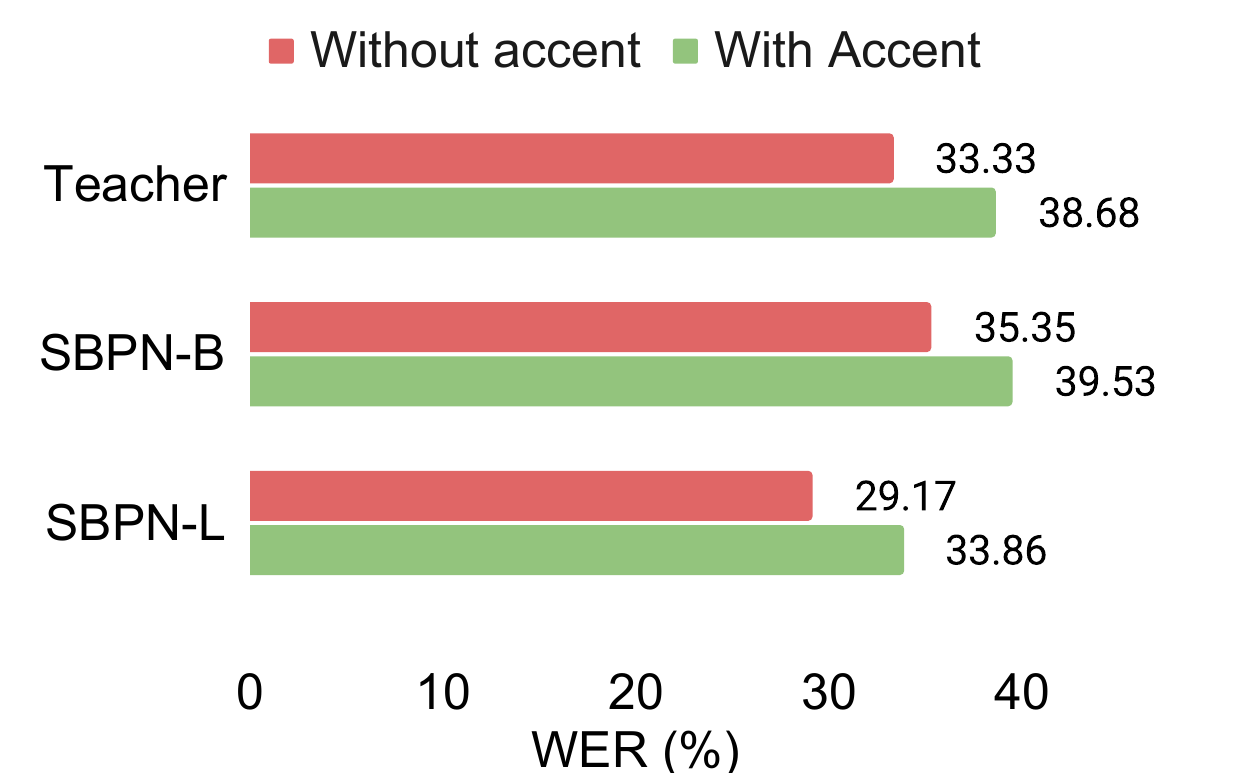}
        \caption{Igbo}
        \label{fig:accent_wer_igbo}
    \end{subfigure}
    \caption{Average WER (\%) of SBPN and monolingual Teacher models on the FLEURS test sets before and after removing diacritical marks from a) Yorùbá and b) Igbo predicted texts. The teacher models are the monolingual baselines.}
    \label{fig:accent_wer}
\end{figure}

We examined the effect of diacritical marks on the accuracy of model predictions on the FLEURS test sets for the two Nigerian languages, Yorùbá and Igbo, which have diacritical marks in their orthography. For Yorùbá language, it can be deduced from Fig.~\ref{fig:accent_wer_yor} that the teacher model's high WER is mainly due to its inability to predict the correct mark, leading to a 110~\% relative increase in WER from when diacritics are removed from predictions to when diacritics are retained after prediction. This gap was reduced through knowledge distillation and self improvement training for SBPN-Base and SBPN-Large by 69~\% and 76~\% respectively. An examination of the same setup using the Common Voice test sets (not shown in the plots) indicates only a relative increase in WER of 27~\% and 47~\% from when diacritical marks are removed to when they are retained. In summary, more work still needs to be done to reduce the WER gap between accented and unaccented predictions in Yorùbá language.
For the results of the Igbo language in Fig.~\ref{fig:accent_wer_igbo}, the gap is not as large as that of the Yorùbá language, perhaps because marks are often omitted in standard Igbo writing. Also, the WERs are already relatively higher, so diacritics may not be the largest driver of WER improvement on the Igbo language, as pointed out in Section~\ref{sec:average_tachers}. 

\subsection{Evaluation on language identification (LID)}

We examine SBPN's LID prediction accuracy using the speech samples in the test sets. The results are compared with the other audio-based ECAPA-TDNN and text-based Afro-LID models. Table~\ref{tab:audio_based_id} shows the micro averaged $F_1$ score for each model. In general, SBPN models are on par with the baseline LID models on Igbo, Yorùbá, and Hausa languages. They also perform better in identifying Nigerian English and Nigerian Pidgin utterances.

\begin{table}[t]
\centering
\caption{Audio-based and text-based language identification. For SBPN models, the language tag was selected as the class label. Table shows micro-averaged $F_1$ score for each language.}
\label{tab:audio_based_id}
\begin{tabular}{ l  c  c  c  c }
\toprule
Language & ECAPA-TDNN & AfroLID &  &  \\
{} &  (audio-based) &  (Text-based) & SBPN-Base & SBPN-Large \\
\midrule
en-ng & 20.23 & - & 100.0 & 100.00 \\

yo & 96.03 & 100.00 & 100.00 & 100.00 \\

ha & 97.42 & 99.84 & 99.68 & 100.00 \\

pcm & - & 44.39 & 97.31 & 96.52 \\

ig & - & 100 & 99.69 & 100.00 \\
\bottomrule

\end{tabular}

\end{table}

\section{Conclusion and Future Work}
\label{sec:future}
We introduced the Sometin Beta Pass Notin (SBPN) family of multilingual ASR models (SBPN-Base with 120M parameters and SBPN-Large with 600M parameters) targeting five Nigerian languages: Hausa, Yorùbá, Igbo, Nigerian English, and Nigerian Pidgin. Evaluation results demonstrate that SBPN models consistently outperform existing monolingual and multilingual baselines across both read speech and conversational speech benchmarks. Future work will focus on expanding language coverage to additional Nigerian languages and developing targeted methods to improve diacritic prediction accuracy for tonal languages such as Yorùbá and Igbo.

\section*{Acknowledgment}
The author appreciates Amina Mardiyyah Rufai for proofreading the manuscript.

\bibliographystyle{ieeetr}
\bibliography{refs}

\end{document}